\documentclass[letterpaper, 10 pt, conference]{ieeeconf}  

\IEEEoverridecommandlockouts                              

\usepackage{times}
\usepackage{amsmath}
\usepackage{amssymb}
\usepackage{mathtools}
\usepackage{graphicx}
\usepackage{multirow}
\usepackage{makecell}
\usepackage{booktabs}
\usepackage{algorithm}
\usepackage{algorithmic}
\usepackage{url}
\usepackage{hyperref}
\usepackage{caption}
\usepackage{xcolor}
\usepackage{subcaption}
\usepackage{float}
\usepackage{balance}

\def \MethodAcronym {ViTaMIn-B}
\def \tasknumber {4}

\title{\LARGE \bf \MethodAcronym{}: A Reliable and Efficient Visuo-Tactile Bimanual Manipulation Interface}

\author{\authorblockN{Chuanyu Li$^{*,1}$, Chaoyi Liu$^{*,1}$, Daotan Wang$^{1}$, Shuyu Zhang$^{4}$ \\ Lusong Li$^{3}$, Zecui Zeng$^{3}$, Fangchen Liu$^{2}$, Jing Xu$^{\dagger,1}$, Rui Chen$^{\dagger,1}$}
${}^{1}$Tsinghua University, ${}^{2}$University of California, Berkeley\\
${}^{3}$JD Explore Academy, ${}^{4}$The Hong Kong Polytechnic University\\
$*$ Equal contribution, 
$\dagger$ Equal advising
\\
\url{https://chuanyune.github.io/ViTaMIn-B_page/}}

\begin{document}
\bstctlcite{setting}

\let\oldtwocolumn\twocolumn
\renewcommand\twocolumn[1][]{%
  \oldtwocolumn[{#1}{
    \vspace{-8mm} 
    \begin{center}
      \includegraphics[width=\textwidth]{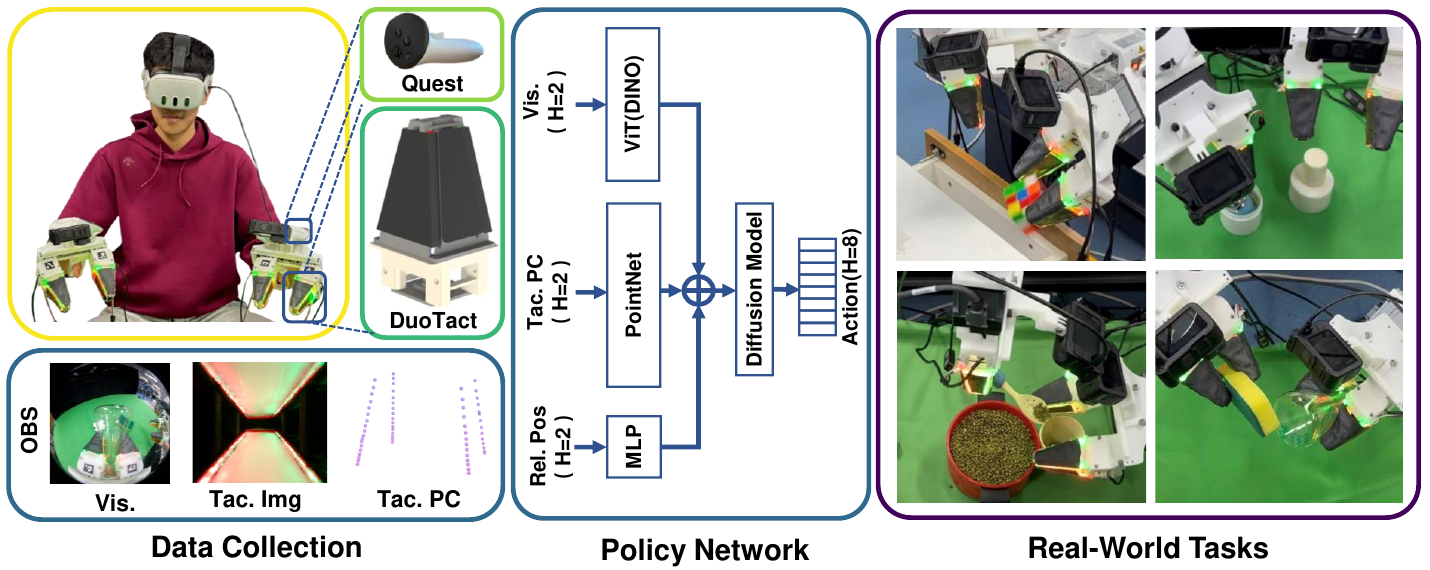}
      \vspace{-5mm}
      \captionof{figure}{We present \MethodAcronym, a  handheld system specifically designed for collecting visuo-tactile demonstrations in contact-rich bimanual manipulation tasks.
The left panel illustrates the data-collection setup and multimodal observations, the middle panel shows the policy network used for learning from demonstrations, and the right panel depicts a wide variety of manipulation tasks performed using the proposed system.}
      \label{fig:teaser}
    \end{center}
  }]
}

\maketitle
\thispagestyle{empty}
\pagestyle{empty}

\bstctlcite{IEEEexample:BSTcontrol}

\begin{abstract}
Handheld devices have opened up unprecedented opportunities to collect large-scale, high-quality demonstrations efficiently. However, existing systems often lack robust tactile sensing or reliable pose tracking to handle complex interaction scenarios, especially for bimanual and contact-rich tasks. In this work, we propose \MethodAcronym{}, a more capable and efficient handheld data collection system for such tasks. 
We first design DuoTact, a novel compliant visuo-tactile sensor built with a flexible frame to withstand large contact forces during manipulation while capturing high-resolution contact geometry. To enhance the cross-sensor generalizability, we propose reconstructing the sensor's global deformation as a 3D point cloud and using it as the policy input.
We further develop a robust, unified 6-DoF bimanual pose acquisition process using Meta Quest controllers, which eliminates the trajectory drift issue in common SLAM-based methods. Comprehensive user studies confirm the efficiency and high usability of \MethodAcronym{} among novice and expert operators. Furthermore, experiments on four bimanual manipulation tasks demonstrate its superior task performance relative to existing systems.
\end{abstract}

\section{Introduction}
Learning from demonstrations has become a cornerstone of robot manipulation policy learning. Existing demonstration collection approaches are typically categorized as follows: Teleoperation systems (e.g., \cite{aldaco2024aloha, zhao2023learning}) offer the benefit of highly accurate robot trajectories, but they are inherently cost-intensive, as each system necessitates a dedicated robot. Conversely, Simulation platforms \cite{gu2023maniskill2} enable the generation of large-scale datasets; however, the persistent sim-to-real gap largely limits their real-world applicability, especially in contact-rich scenarios \cite{josifovski2022analysis, chensim2real, tao2025maniskill}.

Against this backdrop, handheld systems \cite{chi2024universal, shafiullah2023bringing} have recently demonstrated a promising capability, enabling cost-effective data collection decoupled from the physical robot. Nevertheless, current handheld systems face significant limitations in challenging manipulation tasks, specifically those that are more fine-grained, involve bimanual interactions, or require rich contact with objects. These tasks necessitate highly accurate gripper tracking to produce high-quality trajectories and often rely on touch sensing as a crucial complementary modality for proper object interaction.

However, existing handheld systems fall short of these requirements on several fronts:
(1) SLAM-based two-gripper pose estimation \cite{liu2024forcemimic, li2024touchinthewild, chi2024universal, liu2025vitamin} is fragile and prone to trajectory drift when scene changes occur during manipulation, especially for bimanual contact-rich manipulation tasks.
(2) Current visuotactile sensors are frequently too rigid for compliant grasping or structurally inadequate for the substantial forces encountered in bimanual tasks \cite{alltact, li2024touchinthewild}. (3) Policies trained directly on raw touch sensor images exhibit unsatisfactory generalization across different sensor instances due to manufacturing variations.

As shown in Fig. 1, we present \textbf{\MethodAcronym{}}, a reliable and efficient handheld system designed to collect bimanual vision-tactile demonstrations, which effectively addresses these challenges. Our contributions are threefold:
\begin{itemize}
    \item \textbf{DuoTact Visuotactile Sensor}: We introduce DuoTact, a novel visuotactile sensor featuring a flexible frame and two contact faces. This design enables compliant grasping for better object adaptation and enhanced grasp stability, while structurally withstanding forces exceeding 10 N.
    \item \textbf{Generalizable Tactile Representation}: To significantly improve the policy's cross-sensor generalizability, we propose reconstructing the global deformation into a point cloud representation. We then design a policy network specifically to integrate this tactile modality alongside visual and proprioceptive inputs.
    \item \textbf{Robust Bimanual Pose Tracking}: We employ Meta Quest 3 controllers to acquire the poses of both grippers within a unified coordinate system. This approach eliminates the trajectory drift issue experienced by SLAM-based methods under scene changes. Furthermore, we develop per-modality latency compensation to synchronize all sensing modalities within a tight 10 ms window.
\end{itemize}

We evaluate the effectiveness of \MethodAcronym{} on \tasknumber{} bimanual manipulation tasks. Experiment results demonstrate that tactile sensing substantially improves success rates over vision-only baselines, with our novel visuotactile sensor and point-cloud representation yielding superior robustness and generalizability. Ablation studies further validate that temporal alignment and Quest-based pose acquisition are essential components for robust performance. User studies confirm that even novice users can collect high-quality data efficiently. To facilitate future research, the design files and manufacturing process of the entire system will be released.

\section{Related Work}

\subsection{Visuotactile Sensor}
Compared with piezoelectric tactile sensor~\cite{huang20243d}, visuotactile sensors offer significant advantages in high spatial resolution and low manufacturing cost~\cite{li2024vision}. However, existing visuotactile sensors mostly are used in rigid fingers, limiting the grasp stability.
GelSight Fin Ray and GelSight Baby Fin Ray~\cite{9762175, liu2023gelsight} incorporate tactile sensing into a compliant gripper. But the rigid backing used in these designs limits compliance to the contact face and prevents global compliance and omni-directional tactile sensing which are critical for manipulation in dynamic scenarios. PneuGelSight~\cite{zhang2025pneugelsight} uses a pneumatically actuated soft finger with an embedded in-body camera for single-camera proprioception and contact. But it requires a compressed air supply, making it impractical for a low-cost handheld interface. Furthermore, it can only bend along a fixed trajectory, resulting in relatively poor compliance compared to passive deformation.
AllTact~\cite{alltact} is able to deform globally and senses multi-directional contact, yet its unframed, highly compliant bodies struggle to sustain large loads and to stably grasp heavy objects. 
In this work, \textbf{DuoTact} is specifically designed to be both globally compliant and robust enough to withstand large contact forces.

\subsection{Data Collection System for Robot Manipulation}
Collecting high-quality demonstrations is crucial for robot learning~\cite{chi2023diffusion}.  While teleoperation with physical robots provides accurate trajectories, it is costly and embodiment-specific. Handheld, robot-free devices~\cite{chi2024universal} offer a low-cost, flexible alternative and have garnered growing attention.

The original UMI~\cite{chi2024universal} introduces a portable gripper with SLAM-based pose acquisition.
ForceMimic~\cite{liu2024forcemimic} incorporates 6D F/T sensing for contact-rich tasks, though its force signals remain low-dimensional. Touch in the Wild~\cite{li2024touchinthewild} employs piezoresistive sensors, while ViTaMIn~\cite{liu2025vitamin} uses the AllTact visuotactile sensor. However, these works rely on vision-based SLAM for pose tracking, which is prone to failure when the scene changes significantly during manipulation (e.g., cabinet opening).
In this work, we integrate VR controllers into a bimanual handheld system, enabling robust pose tracking independent of environmental dynamics. Concurrent work exUMI~\cite{yu2024exumi} also leverages Meta Quest 3 for tracking. The key distinctions are that we introduce {DuoTact}, a novel compliant visuotactile sensor, and propose using reconstructed deformation point clouds as policy input. In contrast, exUMI relies on raw tactile images from 9DTact~\cite{lin20249dtact}, which senses contact only on the grasping face and limits cross-sensor generalization.

\subsection{Bimanual Robot Manipulation}
Bimanual manipulation enables robots to perform complex tasks requiring coordination between two arms~\cite{SMITH20121340}. Learning-based approaches have shown promise for for such tasks~\cite{zhao2023learning}, with recent efforts achieving success in cloth folding and object assembly. Nonetheless, acquiring high-quality bimanual demonstrations remains difficult due to the requirements for precise coordination and synchronized multimodal sensing~\cite{chitturi2021task}.

Systems like ALOHA~\cite{aldaco2024aloha} and Mobile ALOHA~\cite{fu2024mobile} have advanced bimanual teleoperation via leader-follower configurations, yielding impressive performance on intricate manipulation tasks. However, these approaches necessitate physical robot access during data collection and incur substantial hardware costs. In contrast, exiting handheld systems have mainly targeted single-arm manipulation, leaving a critical gap in robot-agnostic bimanual demonstration collection with integrated multimodal sensing. Our work fills this gap by introducing a bimanual handheld system that captures both grippers' poses in a unified coordinate frame while synchronizing vision and tactile sensing signals—enabling efficient collection of bimanual visuotactile demonstrations without any robot hardware.

\section{Duo-Material Visuotactile Sensor}

In this work, we present DuoTact, which consists of a flexible TPU frame and two PVC-silicone contact faces. The TPU frame enable the sensor to handle larger contact forces, while the PVC-silicon layer enable to sense the shape of the entire sensor and contact geometry details.  

\subsection{Structural Composition}

\begin{figure}[t]
    \centering
    \includegraphics[width=0.4\textwidth]{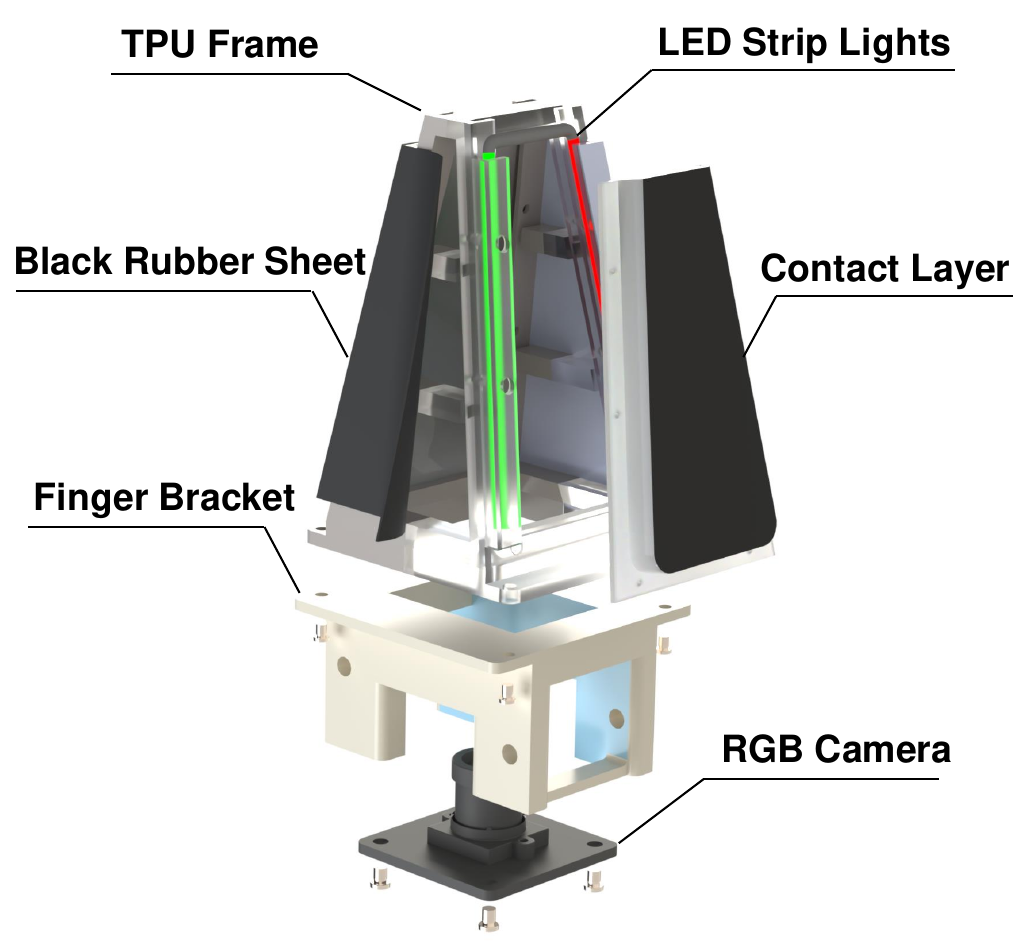}
    \vspace{-3mm}
    \caption{Exploded view of DuoTact structure.}
    \label{fig:sensor_bomb}

\end{figure}

The structure of DuoTact is illustrated in Fig.~\ref{fig:sensor_bomb}:

(1) \textit{TPU Frame}: The flexible frame serves as the structural support for the sensor. It is designed to deform globally to adapt to object shapes during manipulation and return to its original configuration when contact is released.

(2) \textit{Contact Layer}: The multi-layer structure consists of: A 1.6-mm-thick transparent PVC film to support the contact layer, allowing it to bend without collapse; A transparent silicone gel base deforming to reveal local contact conditions; A reflective layer scattering lights to highlight contact details; A black coating overlay to block ambient light interference. 

(3) \textit{Black Rubber Sheet}: Black rubber sheets are heat-sealed on the two sides of the TPU frame to prevent ambient light from interfering with internal image capture.

(4) \textit{LED Strip Lights}: To capture the contact geometry details on the surface, programmable LED strip lights (400 mm × 2.7 mm) are integrated within the TPU frame. Moreover, the illumination can enhance the visibility of the contact layer’s edges in the image, facilitating more accurate point cloud reconstruction.

(5) \textit{Finger Bracket}: This component secures the TPU frame via grooves and serves as the mounting point for the RGB camera beneath it.

(6) \textit{RGB Camera}: An RGB camera that operates at 640×480 resolution, 30 fps is positioned to capture the internal TPU frame and internally illuminated contact layers.

\subsection{Fabrication Process}

\begin{figure}[t]
    \centering
    \includegraphics[width=0.5\textwidth]{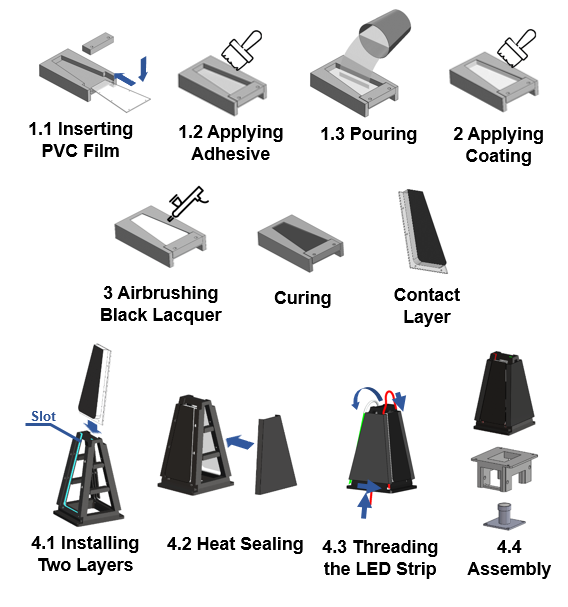}
    \vspace{-10mm}
    \caption{Diagram of the fabrication process for DuoTact}
    \label{fig:fab}
\end{figure}

The fabrication process of DuoTact is illustrated in Fig.~\ref{fig:fab}:

(1) A PVC film is inserted into the mold cavity and coated with a transparent silicone adhesive. Subsequently, 10g of transparent silicone gel (Wacker Elastosil® RT 601, A:B = 9:1 by weight) is poured into the mold and cured at 60°C for 30 minutes.

(2) A reflective coating, consisting of Posilicone Translucent silicone (A:B = 1:1 by weight) and white pigment (2:0.1 weight ratio), is applied onto the cured transparent silicone surface. This layer is then cured at 60°C for 20 minutes.

(3) A black coating, consisting of Novocs Matte matting agent, Ecoflex 00-10 (A:B = 1:1 by weight), and black pigment (26:6:1 weight ratio), is uniformly airbrushed over the reflective layer and dried at 60°C for 20 minutes.

(4) For final assembly, the two fabricated contact layers are slid into slots on the TPU frame. A black rubber sheet is then attached to the frame's exterior via heat sealing. Subsequently, the LED strip light is threaded through designated slots on the frame. Finally, the RGB camera and the TPU frame assembly are mounted onto the finger bracket using screws and nuts.

\subsection{Global Deformation Reconstruction}

\begin{figure}[t]
    \centering
    \includegraphics[width=0.5\textwidth]{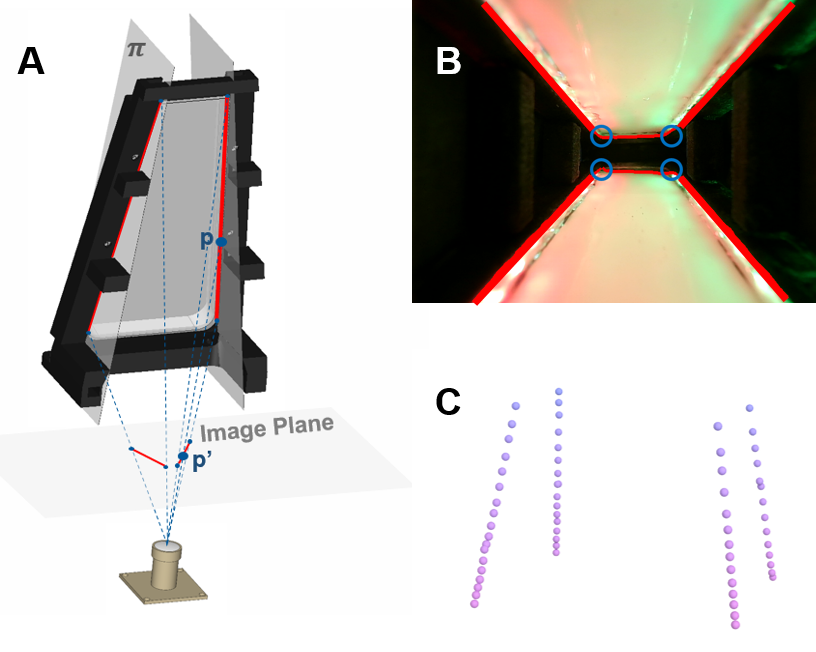}
    \vspace{-8mm}
    \caption{Principle and result diagram of point cloud reconstruction. In Figure A, red lines represent inner edges of the frame. In Figure B, red lines represent detected inner edges in the photo, while the blue circles denote the corner points on the sensor top. Figure C shows the reconstruction effect.}
    \label{fig:reconstruction_principle}
\end{figure}

The captured raw tactile images may vary across sensors due to manufacturing imperfections, which deteriorates the cross-sensor generalizability of learned policies. Therefore, we propose to reconstruct the globally deformed point cloud of the sensor from a single image by utilizing the edge features and spatial constraints present in the image. This point cloud is then used as the input to the policy network. The reconstruction principle is shown in  Fig. ~\ref{fig:reconstruction_principle}A.

Based on the pinhole camera model, the pixel coordinate and spatial coordinate of a point are related as:

\begin{equation}
    z\mathbf{p}'=\mathbf{K}\mathbf{p}
\end{equation}
where $\mathbf{K}\in \mathbb{R}^{3\times 3}$ is the intrinsic matrix of the camera, obtained through calibration, $\mathbf{p}'$ and $\mathbf{p}$ represent the pixel coordinate in the image and the 3D coordinate in the camera coordinate system, respectively:

\begin{equation}
    \mathbf{p}'=(u,v,1)^\mathrm{T}, \mathbf{p}=(x,y,z)^\mathrm{T}
\end{equation}

where z is the unknown variable to be solved.
To solve for $\mathbf{p}$, we first extract the pixel coordinates of the edge lines of the TPU frame using binarization and edge detection. Since the flexible frame only deforms in one direction, the detected inner edge of the frame remains a planar curve. Consequently, point $\mathbf{p}$ on this edge must satisfy the plane equation:

\begin{equation}
    \pi:\mathbf{n}^\mathrm{T}\mathbf{p}=b
\end{equation}
where $\mathbf{n}$ and $b$ are the plane's normal vector and scalar parameter, respectively.

These parameters are initially defined by the sensor's geometry in its CAD coordinate system. Computing $\mathbf{p}$ requires this plane equation to be expressed in the camera coordinate system. We therefore calibrate the transformation using the Perspective-n-Point (PnP) method, based on known marker coordinates in the CAD frame and their corresponding pixel coordinates in the image. In this work, we use a polygon fitting algorithm to fit the edge of each contact layer into trapezoids and extract the four corner points on the top of the sensor (see Fig.~\ref{fig:reconstruction_principle}B) as markers for the PnP method.

After transformation calibration, we can determine the 3D spatial coordinates of each point along the frame edge from a single image, thereby reconstructing the globally deformed point cloud of the gripper as shown in Fig. ~\ref{fig:reconstruction_principle}C.

\section{Demonstration Collection System}

\subsection{System Overview}
\begin{figure}[t]
    \centering
    \includegraphics[width=0.45\textwidth]{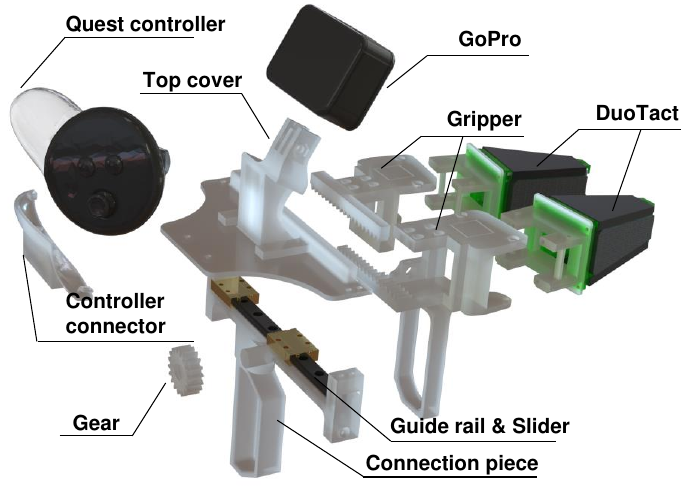}
    \vspace{-3mm}
    \caption{Hardware composition of the \MethodAcronym{} handheld device. The system integrates a Quest controller for bimanual pose tracking, DuoTact sensors for tactile sensing, and modular mechanical parts for stable and ergonomic bimanual operation.}
    \label{fig:data_collection_device}
\end{figure}

\textbf{ViTaMIn-B} is a system developed for bimanual visuo-tactile data collection. As shown in Fig.~\ref{fig:data_collection_device}, the system integrates a GoPro Hero 10 camera for vision observation, Meta Quest 3 controllers for robust 6-DoF bimanual pose acquisition, and two \textbf{DuoTact} sensors for tactile sensing. Gripper width with a maximum span of $8\mathrm{cm}$ is computed by detecting ArUco markers on the gripper. 
As bimanual manipulation demonstration collection occupies both hands, a foot pedal is used to trigger the start and end of recording, enabling efficient single-operator data collection.

Compared with ViTaMIn~\cite{liu2025vitamin}, ViTaMIn-B introduces several key hardware and system-level enhancements.
First, the novel visuotactile sensors (DuoTact) are developed to produce clearer tactile signals across diverse contact scenarios and better contact support.
Second, we replace the SLAM-based tracking with the Meta Quest 3, providing accurate, real-time 6-DoF poses for both handheld devices within a unified coordinate frame and eliminating trajectory loss under large scene variations.
Third, the mechanical structure is orignally designed for improved ergonomics and reduced weight by removing onboard computing (e.g., Raspberry Pi) and interfacing all sensors directly with the host computer. This streamlined architecture significantly enhances usability and reduces operator fatigue during extended data-collection sessions.
Finally, all sensing modalities are latency-calibrated and synchronized to ensure precise spatiotemporal alignment.
Compared with UMI~\cite{chi2024universal}, our trajectory acquisition method is inherently robust to scene variations. Moreover, the overall hardware is lighter and mechanically simpler, featuring fewer transmission components, which improves durability and ease of operation.

\subsection{Multi-Modality Synchonization}
Our collected data consists of two primary sources: images captured by six cameras (one visual camera and two tactile cameras per hand) and pose data obtained from the Meta Quest 3 controllers. To achieve precise temporal alignment, we measure and compensate for the transmission latency of each data modality.

For the Quest pose data, we establish the latency by exchanging synchronization signals between the Quest and the computer. The computer records the signal transmission time $t_{\text{actual}}$ and the reception time $t_{\text{rec}}$, from which we compute the latency $t_{\text{latency}} = t_{\text{rec}} - t_{\text{actual}}$. 

For the camera data, we employ an ArUco marker-based approach. The host computer displays a time-varying ArUco marker on the screen with a known display time $t_{\text{actual}}$. Each camera captures this marker. When the computer receives the camera frames, it records the reception time $t_{\text{rec}}$ for each frame. By detecting the ArUco marker in the received frames and matching it with the corresponding display time, we determine the frame-specific latency $t_{\text{latency}}$ for each camera. This per-modality latency measurement enables us to reconstruct the true capture timestamps for all data streams, ensuring precise temporal alignment across all sensing modalities.
Visual and tactile cameras operate at $30\,\mathrm{Hz}$, while the Quest pose stream runs at $72\,\mathrm{Hz}$; after temporal correction, we interpolate the Quest poses onto the visual frame timestamps so that all modalities share a common time base.

\subsection{Policy Network Design}
For the tactile modality, we represent each tactile frame as a point cloud (256 points) and encode it with a \emph{shared} PointNet~\cite{qi2017pointnet}, yielding a $768$-D feature per sensor (weights are tied across sensors).
Visual observations are encoded by a DINO-pretrained ViT-B/16~\cite{dosovitskiy2020vit} into a $768$-D image embedding.
Proprioceptive states are embedded by a two-layer MLP into a $256$-D vector.
We concatenate all embeddings and feed them to a diffusion-policy network~\cite{ho2020denoising} trained with DDIM~\cite{song2020denoising}.

\subsection{Transformation Calibration}

During  data collection, we use a Quest controller fixed on the handheld device to  record the trajectory of the gripper. The trajectory will be recorded as a sequence of transform from the controller frame (Q) to the system's world frame (W), i.e. ${}^{W}\mathbf{T}_{Q_i}(i=1,2,3...)$. Notably, both the controller frame and the world frame provided by Quest are left-handed frames. We have to convert them to equivalent right-handed frames at first.

However, during deployment, we use the the transform of the robot end effector (EE) pose relative to the previous moment, i.e. ${}^{EE_{i+1}}\mathbf{T}_{EE_i}$, to control the robotic arm. Therefore, we need to calibrate the hand-eye transform between the Quest Controller and the EE, i.e. ${}^{Q}\mathbf{T}_{EE}$.

\begin{figure}[t]
    \centering
    \includegraphics[width=0.4\textwidth]{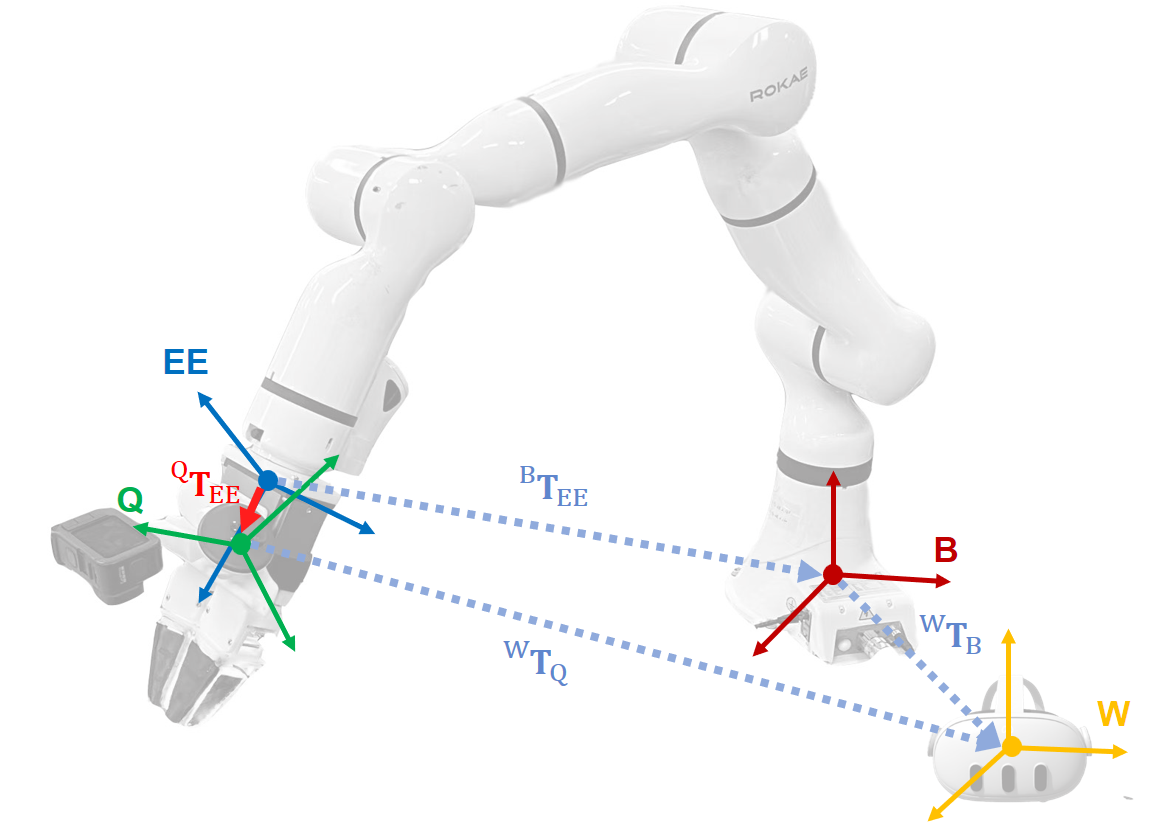}
    \caption{The relationship between frames used in the transformation calibration process. Arrows in the picture denote transforms between frames, where the red one denotes the desired hand-eye transform ${}^{Q}T_{EE}$.}
    \label{fig:hand-eye}
    \vspace{-2mm}
\end{figure}

During  hand-eye calibration, we mount the Quest controller on the end of the  robotic arm (see Fig.~\ref{fig:hand-eye}) and move the robotic arm to $n$ sampling poses ($n=10$ in this work), recording the corresponding ${}^{W}\mathbf{T}_{Q_{k}}$ and ${}^{B}\mathbf{T}_{EE_{k}}(k=1,2,...,n)$, where $B$ denotes the base frame of the robotic arm. For each of them, the following relationship holds:

\begin{equation}
    {}^{W}\mathbf{T}_{Q_{k}} {}^{Q}\mathbf{T}_{EE} = {}^{W}\mathbf{T}_B {}^{B}\mathbf{T}_{EE_k}
\end{equation}

Thus, the desired hand-eye transformation ${}^{Q}\mathbf{T}_{EE}$ can be solved~\cite{doi:10.1177/027836499501400301}. After data collection, we will calculate the corresponding ${}^{EE_{i+1}}\mathbf{T}_{EE_i}$ based on the collected ${}^{W}\mathbf{T}_{Q_i}$ and ${}^{Q}\mathbf{T}_{EE}$, and use it as training data.

\section{Experiments}

\subsection{Experimental Setup and Tasks}

\begin{figure*}[t]
    \centering
    \includegraphics[width=\textwidth, height=0.45\textheight]{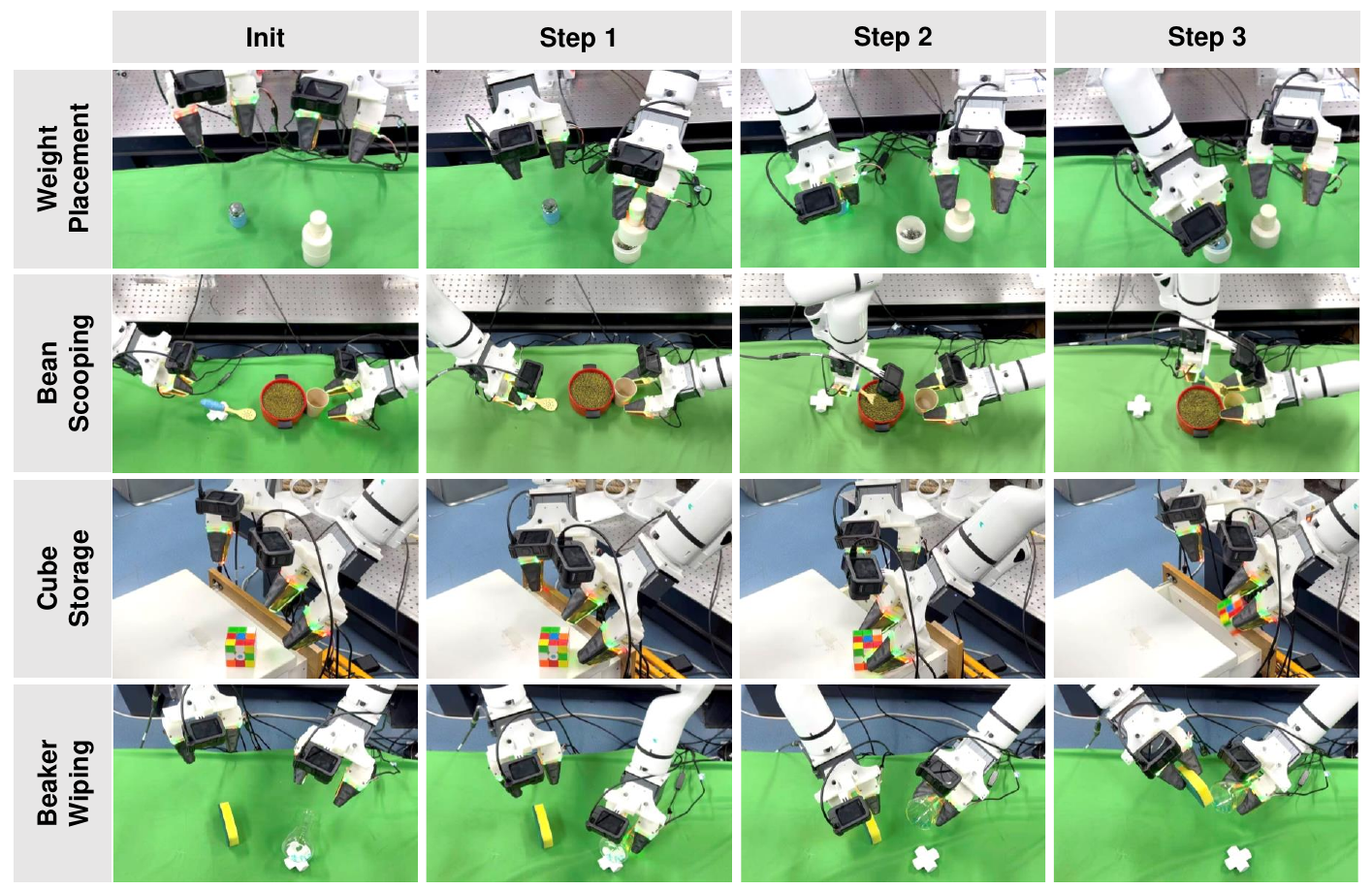}
    \caption{Bimanual manipulation tasks, including Weight Placement, Bean Scooping, Cube Storage, and Beaker Wiping. Each row illustrates sequential steps from initialization to task completion.}
    \label{fig:tasks}
    \vspace{-1mm}
\end{figure*}

We evaluate the effectiveness of the proposed data collection system on \tasknumber{} bimanual manipulation tasks requiring varied levels of coordination. All demonstrations are collected using our system. Dataset statistics are summarized in Table~\ref{tab:data_statistics}, and detailed task procedures are illustrated in Fig.~\ref{fig:tasks}.

\begin{figure*}[htbp]
    \centering
    \includegraphics[width=\textwidth]{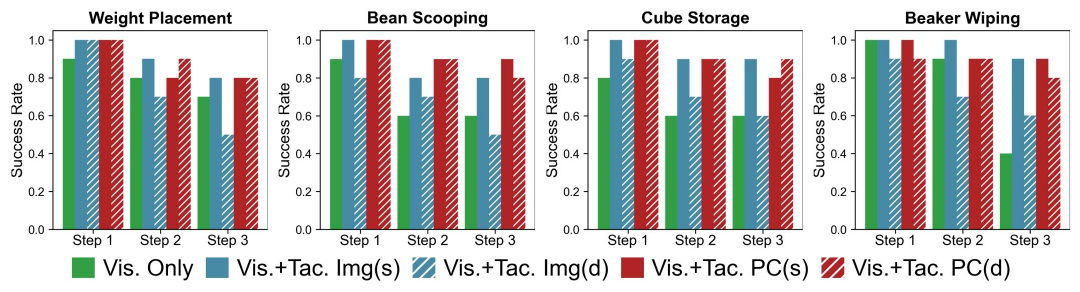}
    \vspace{-7mm}
    \caption{Step-wise success rates across 4 bimanual manipulation tasks with different input representations.
    Here, (s) denotes experiments where the same tactile sensors are used for both data collection and deployment, while (d) indicates using different sensors across the two stages.}
    \label{fig:task_results}
    \vspace{-6mm}
\end{figure*}

\begin{table}[htbp]
    \centering
    \caption{Dataset statistics for the four bimanual manipulation tasks, including the number of demonstrations and the average sequence length (in frames) per task.}
    \label{tab:data_statistics}
    \renewcommand{\arraystretch}{0.85}
    \small
    \begin{tabular}{lccc}
        \toprule
        \textbf{Tasks} & \textbf{Demos} & \textbf{Avg. Frames} \\
        \midrule
        Weight Placement & 123 & 337 \\
        Bean Scooping & 345 & 383 \\
        Cube Storage & 195 & 286 \\
        Beaker Wiping & 181 & 437 \\
        \bottomrule
    \end{tabular}
    \renewcommand{\arraystretch}{1}
\end{table}

\textbf{Task 1: Weight Placement}  The objective is to place a weight block into a box. This task consists of three steps: (1) the left hand opens the lid, (2) the right hand grasps a 500 g weight block, and (3) the right hand places the block into the box. The key challenge lies in the precise alignment and insertion required due to the tight 5 mm clearance between the block and the box opening.
\textbf{This task primarily evaluates the grippers' ability to stably manipulate relatively heavy objects under precise alignment constraints.}
Both the weight block and box are randomly positioned within a $40\mathrm{cm}\times40\mathrm{cm}$ workspace.

\textbf{Task 2: Bean Scooping} The objective is to scoop beans from a pot and transfer them to a cup. The left hand grasps a spoon, reaches into a pot, scoops out mung beans, and pours them into a cup.
\textbf{This task mainly tests coordinated bimanual manipulation, where both arms must collaborate to transport and accurately deposit granular materials.}
The initial robot pose varies within a range, and the spoon and cup are randomly positioned within a $30\mathrm{cm}\times30\mathrm{cm}$ workspace. 

\textbf{Task 3: Cube Storage} The objective is to store a Rubik's cube in a drawer. One hand opens the drawer while the other hand grasps the Rubik's cube and places it into the opened drawer.
\textbf{This task evaluates the system's ability to manipulate articulated objects while coordinating the motion of both hands.}
The drawer position remains fixed, while the Rubik's cube is randomly positioned within a $20\mathrm{cm}\times20\mathrm{cm}$ region.

\textbf{Task 4: Beaker Wiping} The objective is to clean markings from the bottom of a cup. The left hand grasps a transparent cup while the right hand grasps a sponge. Both hands coordinate to precisely wipe off the markings at the bottom of the cup.
\textbf{This task is designed to assess performance on contact-rich bimanual manipulation, requiring sustained visuo-tactile feedback during fine-grained wiping motions.}
The cup and sponge are randomly positioned within a $45\mathrm{cm}\times40\mathrm{cm}$ workspace.

\subsection{Bimanual Task Evaluation}

We compare three input representations: vision-only, vision+tactile image, and vision+tactile point cloud (PC). Each is evaluated on all the \tasknumber{} tasks, with success rates measured for individual sub-steps. 
To evaluate the cross-sensor generalizablity,  we design a comparison experiment: (1) using consistent gripper installation sequences across data collection and deployment phases, and (2) alternating the installation sequences of the grippers between data collection and deployment phases. 

Results are shown in Fig.~\ref{fig:task_results}. The results demonstrate that incorporating tactile sensing substantially improves task success rates, particularly for tasks requiring precise force control and object manipulation (e.g., Step 3 in Bean Scooping and Beaker Wiping). Both tactile-enhanced approaches show consistent improvements across all sub-steps compared to the vision-only baseline. Notably, the tactile point cloud representation demonstrates strong generalizability, maintaining high performance even when a different sensor is used for deployment than for data collection. This suggests the point cloud is a more robust representation to sensor-to-sensor variations compared to raw tactile images.

\begin{figure}[htbp]
    \centering
    \includegraphics[width=0.48\textwidth]{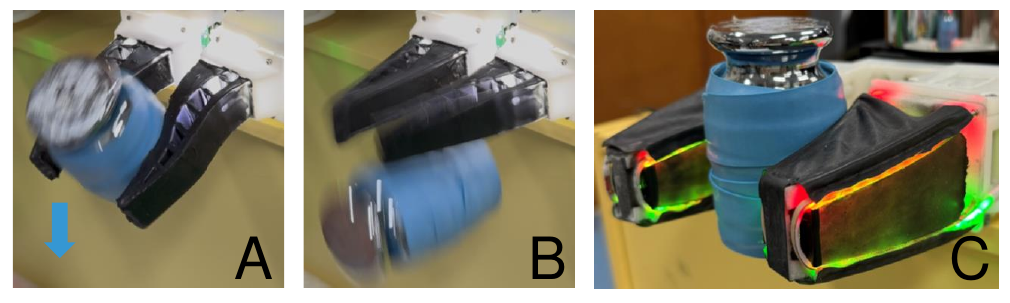}
    \vspace{-7mm}
    \caption{Comparison of gripping performance when grasping a 1\,kg weight. (A–B) The AllTact-based gripper exhibits noticeable slip and cannot maintain a stable grasp under large tangential loads. (C) The proposed DuoTact sensor maintains a stable grasp on the same 1\,kg weight, demonstrating improved support capability.}
    \label{fig:sensor_camparation}
\end{figure}

\subsection{Sensor Comparison}

The AllTact sensor~\cite{alltact} used in ViTaMIn~\cite{liu2025vitamin} is found to be unable to withstand large contact forces, as its design lacks a supporting frame. In ViTaMIn-B, we address this by designing a novel sensor DuoTact with a flexible frame structure, substantially improving its force support capability.

To validate this improvement, we conduct a comparative experiment grasping weights of identical mass using AllTact and DuoTact. In addition to the policy-driven Weight Placement task with a 500 g block, we also design a dedicated static grasping task in which the gripper is commanded to lift and hold a 1 kg weight, as shown in Fig.~\ref{fig:sensor_camparation}, to directly evaluate sensor-level load-bearing performance. The AllTact sensor exhibits clear yielding and fails to hold a 500 g weight. In contrast, DuoTact, due to its frame structure, remained stable even when holding a 1 kg weight.

To further evaluate the impact of sensor hardware, we conduct a cross-sensor generalization experiment on the Weight Placement task. For AllTact, we follow ViTaMIn and use raw tactile images as policy input. The results in Table~\ref{tab:sensor_comparison} demonstrate the clear advantage of DuoTact and our point cloud representation. AllTact exhibits low overall success and further degrades when deployed on a different sensor using raw tactile images. A similar cross-sensor drop is observed for the DuoTact image-based policy (from 0.8 to 0.5). In contrast, our tactile point cloud policy consistently maintains a success rate of 0.8 under both same-sensor and cross-sensor deployment, indicating strong robustness to sensor-to-sensor variations.

\begin{table}[htbp]
\centering
\caption{Cross-Sensor Generalization in Weight Placement}
\label{tab:sensor_comparison}
\renewcommand{\arraystretch}{0.85}
\small
\begin{tabular}{llcc}
\toprule
\textbf{Sensor} & \textbf{Config.} & \textbf{Modality} & \textbf{Success Rate} \\
\midrule
\multirow{2}{*}{AllTact~\cite{alltact}}
  & Same        &  Img   & 0.2 \\
  & Diff.       &  Img   & 0.1 (-0.1) \\ 
\midrule
\multirow{4}{*}{DuoTact}
  & Same        & Img   & 0.8 \\
  & Diff.       & Img   & 0.5 (-0.3) \\
  & Same        & PC    & 0.8 \\
  & Diff.       & PC    & \textbf{0.8} (-0) \\
\bottomrule
\end{tabular}
\renewcommand{\arraystretch}{1}
\end{table}

\subsection{System Component Evaluation}

To validate the key design choices in the system, we conduct ablation studies on two critical components: time alignment and pose tracking method.

\subsubsection{Time Alignment}
To demonstrate the importance of the multi-modality synchronization, we train policies on the Weight Placement task using data with and without time alignment.
The observed latency in each modality is mainly due to transmission delays: from sensor exposure to data reception on the PC, we measure a delay of $0.14\,\mathrm{s}$ for the GoPro camera, $0.08\,\mathrm{s}$ for the tactile cameras, and $0.01\,\mathrm{s}$ for the Quest controller over a wired connection.
Table~\ref{tab:temporal_alignment} presents the comparison results.

\begin{table}[htbp]
    \centering
    \caption{Impact of Time Alignment on Weight Placement}
    \label{tab:temporal_alignment}
    \renewcommand{\arraystretch}{0.85}
    \small
    \begin{tabular}{lccc}
        \toprule
        \textbf{Time Alignment} & \textbf{Step 1} & \textbf{Step 2} & \textbf{Step 3} \\
        \midrule
        Without Alignment & 0.5 & 0.3 & 0.2 \\
        With Alignment & \textbf{0.9} & \textbf{0.8} & \textbf{0.7} \\
        \bottomrule
    \end{tabular}
    \renewcommand{\arraystretch}{1}
\end{table}

The results clearly demonstrate that time alignment  is essential for policy learning. Without proper alignment, the misalignment between visual and tactile observations leads to substantial performance degradation, particularly in steps requiring precise visuo-tactile coordination.

\subsubsection{Pose Tracking Method Comparison}
We compare the VR controller-based pose tracking with the SLAM-based approach used in UMI~\cite{chi2024universal}. We collect demonstrations for the weight placement task using both methods and evaluate the data quality and the performances of trained policies. Table~\ref{tab:pose_tracking} presents the comparison results.

\begin{table}[htbp]
    \centering
    \caption{Pose Tracking Method Comparison on Weight Placement}
    \label{tab:pose_tracking}
    \renewcommand{\arraystretch}{0.85}
    \small
    \begin{tabular}{lccc}
        \toprule
        \textbf{Method} & \makecell{\textbf{Collected}\\\textbf{Demos}} & \makecell{\textbf{Valid}\\\textbf{Rate}} & \makecell{\textbf{Success}\\\textbf{Rate}} \\
        \midrule
        SLAM-based~\cite{chi2024universal, liu2025vitamin} & 150 & 0.16 & 0.1 \\
        VR controller-based & 123 & \textbf{1.0} & \textbf{0.7} \\
        \bottomrule
    \end{tabular}
    \renewcommand{\arraystretch}{1}
\end{table}

The VR controller-based tracking demonstrates superior data quality with higher valid demonstration rates. The SLAM-based method suffers from trajectory loss under large scene variations during manipulation, resulting in incomplete or corrupted demonstrations. Furthermore, Quest 3 enables direct acquisition of both grippers' poses in the same coordinate system, eliminating the need for coordinate transformation computation. The success rate column shows the policy performance when trained with data collected by each method.

\subsection{User Study}
To evaluate the usability of the \MethodAcronym{} system, we conduct a user study involving 5 participants: 3 inexperienced users who have never used the system before and 2 experienced users who are familiar with the system. All participants are asked to collect demonstration data for the same \tasknumber{} tasks described above.
The experimental protocol ensure fair comparison by providing identical task instructions and workspace setup for all participants. Inexperienced users receive a brief 5-minute tutorial on system operation before data collection.

We evaluate: (1) Collection time - the time to collect 50 demonstrations, (2) Success rate - the percentage of valid demonstrations among all attempts, (3) Setup time - the time needed to initialize and prepare the system before data collection, and (4) User satisfaction - subjective ratings from participants on ease of use and overall experience.

The results in Table~\ref{tab:user_study} show that the system achieves excellent usability even for novice users, with minimal performance gap between experienced and inexperienced participants. The high success rates and positive user feedback validate the system's intuitive design and robust operation for efficient data collection.

\begin{table}[htbp]
    \centering
    \caption{User Study Results}
    \label{tab:user_study}
    \renewcommand{\arraystretch}{0.85}
    \small
    \begin{tabular}{lcccc}
        \toprule
        \textbf{Participant} 
        & \makecell{\textbf{Time}\\\textbf{(min)} $\downarrow$} 
        & \makecell{\textbf{Success} \\ $\uparrow$}
        & \makecell{\textbf{Setup}\\\textbf{(min)} $\downarrow$} 
        & \makecell{\textbf{Satis.}\\\textbf{(1-10)} $\uparrow$} \\
        \midrule
        Exp. P1 & 19 & 0.99 & 2 & 10 \\
        Exp. P2 & 20 & 1.0 & 3 & 9 \\
        \midrule
        \textbf{Exp. Mean} & \textbf{19.5} & \textbf{0.995} & \textbf{2.5} & \textbf{9.5} \\
        \midrule
        Inexp. P1 & 26 & 1.00 & 5 & 9 \\
        Inexp. P2 & 28 & 0.98 & 4 & 10 \\
        Inexp. P3 & 26 & 0.97 & 5 & 10 \\
        \midrule
        \textbf{Inexp. Mean} & \textbf{26.6} & \textbf{0.983} & \textbf{4.6} & \textbf{9.5} \\
        \bottomrule
    \end{tabular}
    \renewcommand{\arraystretch}{1}
\end{table}

\section{Conclusion}
In this paper, we present ViTaMIn-B, a  handheld device that successfully addresses key challenges in bimanual, contact-rich demonstration collection. Our design integrates a novel compliant visuotactile sensor (DuoTact) with a robust pose tracking system, enabling the collection of high-fidelity multimodal data. Our experiments show that policies trained on this data, particularly using our proposed tactile point cloud representation, achieve high success rates and robust cross-sensor generalization, significantly outperforming vision-only baselines. This work utilizes vision and tactile observations from both hands. A promising future direction is to incorporate a top-down camera to provide global context for manipulation in larger workspaces. Additionally, while current evaluations focus on tabletop tasks, future work could extend to long-horizon, multi-stage mobile manipulation scenarios.

\balance

\bibliographystyle{IEEEtran}
\bibliography{IEEEabrv, ref}

\end{document}